\documentclass[10pt,twocolumn,letterpaper]{article}

\usepackage{wacv}
\usepackage{times}
\usepackage{epsfig}
\usepackage{graphicx}
\usepackage{amsmath}
\usepackage{amssymb}
\usepackage{bbm}
\usepackage{tabularx}
\usepackage{booktabs}
\usepackage{multirow}
\usepackage{xcolor}
\usepackage{setspace}
\usepackage{enumitem}
\usepackage{authblk}
\usepackage[accsupp]{axessibility}

\newcommand*{\affaddr}[1]{#1} 
\newcommand*{\affmark}[1][*]{\textsuperscript{#1}}
\newcommand*{\email}[1]{\texttt{#1}}
\setlist[itemize]{topsep=0pt, partopsep=0pt, parsep=0pt, leftmargin=1em}

\def\wacvPaperID{837}

\wacvfinalcopy 

\ifwacvfinal
\fi

\ifwacvfinal
\usepackage[breaklinks=true,bookmarks=false]{hyperref}
\else
\usepackage[pagebackref=true,breaklinks=true,colorlinks,bookmarks=false]{hyperref}
\fi

\begin{document}

\title{Knowledge-Augmented Contrastive Learning for Abnormality Classification and Localization in Chest X-rays with Radiomics using a Feedback Loop}

\author{%
Yan Han\affmark[1], Chongyan Chen\affmark[1], Ahmed Tewfik\affmark[1], Benjamin Glicksberg\affmark[2], Ying Ding\affmark[1], Yifan Peng\affmark[3]\dag, Zhangyang Wang\affmark[1]\ddag\\
\vspace{-0.3cm}
\affaddr{\affmark[1]The University of Texas at Austin},
\affaddr{\affmark[2]Icahn School of Medicine at Mount Sinai},
\affaddr{\affmark[3]Weill Cornell Medicine}\\
\email{\dag yip4002@med.cornell.edu, \ddag atlaswang@utexas.edu}
}

\maketitle

\ifwacvfinal
\thispagestyle{empty}
\fi

\begin{abstract}
\vspace*{-2em}

Accurate classification and localization of abnormalities in chest X-rays play an important role in clinical diagnosis and treatment planning. Building a highly accurate predictive model for these tasks usually requires a large number of manually annotated labels and pixel regions (bounding boxes) of abnormalities. However, it is expensive to acquire such annotations, especially the bounding boxes. Recently, contrastive learning has shown strong promise in leveraging unlabeled natural images to produce highly generalizable and discriminative features. However, extending its power to the medical image domain is under-explored and highly non-trivial, since medical images are much less amendable to data augmentations. In contrast, their prior knowledge, as well as radiomic features, is often crucial. To bridge this gap, we propose an end-to-end semi-supervised knowledge-augmented contrastive learning framework, that simultaneously performs disease classification and localization tasks. The key knob of our framework is a unique positive sampling approach tailored for the medical images, by seamlessly integrating radiomic features as a knowledge augmentation. 
Specifically, we first apply an image encoder to classify the chest X-rays and to generate the image features. We next leverage Grad-CAM to highlight the crucial (abnormal) regions for chest X-rays (even when unannotated), from which we extract radiomic features. The radiomic features are then passed through another dedicated encoder to act as the positive sample for the image features generated from the same chest X-ray. In this way, our framework constitutes a feedback loop for image and radiomic features to mutually reinforce each other. Their contrasting yields knowledge-augmented representations that are both robust and interpretable. Extensive experiments on the NIH Chest X-ray dataset demonstrate that our approach outperforms existing baselines in both classification and localization tasks. 
\end{abstract}

\section{Introduction}
The chest X-ray is one of the most common radiological examinations for detecting cardiothoracic and pulmonary abnormalities. Due to the demand for accelerating chest X-ray analysis and interpretation along with the overall shortage of radiologists, there has been a surging interest in building automated systems of chest X-ray abnormality classification and localization \cite{rajpurkar2017chexnet}. While the class (i.e., outcomes) labels are important, the localization annotations, or the tightly-bound local regions of images that are most indicative of the pathology, often provide richer information for clinical decision making (either automated or human-based).


Automatic robust image analysis of chest X-rays currently faces many challenges. \underline{First}, recognizing abnormalities in chest X-rays often requires expert radiologists. This process is therefore time-consuming and expensive to generate annotations for chest X-ray data, in particular the localized bounding box region labeling. \underline{Second}, unlike natural images, chest X-rays have very subtle and similar image features. The most indicative features are also very localized. Therefore, chest X-rays are sensitive to distortion and not amendable to typical image data augmentations such as random cropping or color jittering. \underline{Moreover}, in addition to high inter-class variance of abnormalities seen in chest X-rays (i.e., feature differences between different diseases), chest X-rays also have large intra-class variance (i.e., differences in presentation among individuals of the same diseases). The appearance of certain diseases in X-rays are often vague, can overlap with other diagnoses, and can mimic many other benign abnormalities. \underline{Last but not least}, the class distribution of chest X-rays is also highly imbalanced for available datasets.

Recently, contrastive learning has emerged as the front-runner for self-supervised learning, demonstrating superior ability to handle unlabelled data. Popular frameworks include MoCo \cite{he2020momentum, chen2020improved}, SimCLR \cite{chen2020simple, chen2020big}, PIRL \cite{misra2020self} and BYOL \cite{grill2020bootstrap}. They all have achieved prevailing success in natural image machine learning tasks, such as image classification and object detection. Further, contrastive learning appears to be robust for semi-supervised learning when only few labeled data are available \cite{chen2020big}. Recent works also found contrastive learning to be robust to data imbalance \cite{yang2020rethinking,kangexploring}. 

Contrastive learning may offer a promising avenue for learning from the mostly unlabeled chest X-rays, but leveraging it for this task is not straightforward. One most important \textit{technical barrier} is that most contrastive learning frameworks \cite{he2020momentum, chen2020improved, chen2020simple, chen2020big, grill2020bootstrap} critically depend on maximizing the similarity between two ``views", i.e., an anchor and its \textbf{positive sample}, often being generated by applying random data augmentations to the same image. This data augmentation strategy, however, does not easily translate to chest X-rays. In addition, the simultaneous demand for both classification and localization-aware features further complicates the issue. Fortunately, classical chest X-ray analysis has introduced \textbf{radiomic features} \cite{chen2017assessment} as an auxiliary knowledge augmentation. The radiomic features can be considered as a strong \textit{prior}, and therefore can potentially be utilized to guide learning of deep feature extractors. However, the extraction of reliable radiomic features via Pyradiomic\footnote{\url{https://pyradiomic.readthedocs.io/}} tool \cite{van2017computational} heavily depends on the pathology localization -- hence we will run into an intriguing ``chicken-and-egg" problem, when trying to incorporate radiomic features into contrastive learning, whose goal includes learning the localization from unlabeled data.

\begin{figure}
\begin{center}
\includegraphics[width=1\linewidth]{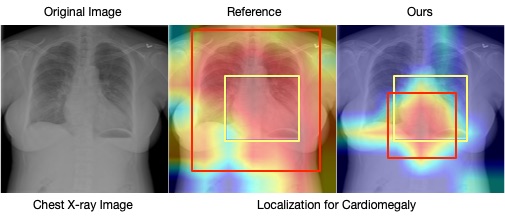}
\end{center}
  \caption{Visualization of heatmaps of chest X-rays with ground-truth bounding box annotations (yellow) and its prediction (red) for localize Cardiomegaly in one test chest X-ray image. The visualization is generated by rendering the final output tensor as heatmaps and overlaying it on the original images. The left image is the original chest X-ray image, the middle is the visualization result by CheXNet \cite{rajpurkar2017chexnet} and the right is our model's attempt. Best viewed in color.}
\label{fig:teaser}
\end{figure}

This paper presents an innovative holistic framework of \textbf{Knowledge-Augmented Contrastive Learning}, which seamlessly integrates radiomic features as the other contrastive knowledge-augmentation for the chest X-ray image. As the \textit{main difference} from existing frameworks, the two ``views" that we contrast now are from two different domain knowledge characterizing the same patient: the chest X-ray image and the radiomic features. Notably, the radiomic features have to be extracted from the learned pathology localizations, which are not readily available. As these features will be dynamically updated, forming a ``feedback loop" during training in which both modalities' learning mutually reinforce each other. The \textit{key enabling technique} to link this feedback loop is a novel module we designed, called \textit{\underline{B}ootstrap \underline{Y}our \underline{O}wn \underline{P}ositive Samples} (\textbf{BYOP}). For an unannotated X-ray image, we utilize Grad-CAM \cite{selvaraju2016grad} to generate the input heatmap from the image modality backbone, which yields the estimated bounding box after thresholding; and we then extract the radiomic features within this estimated bounding box, which becomes the alternative view to contrast with the image view. The usage of radiomic features also adds to the model interpretability. Our contributions are outlined as follows:

\begin{itemize}
\item A brand-new \textbf{framework} dedicated to improving abnormality identification and localization in (mostly unannotated) chest X-rays by knowledge-augmented contrastive learning, which highlights exploiting radiomic features as the auxiliary knowledge augmentation to contrast with the images, given the inability to perform classical image data augmentation. 
\item An innovative \textbf{technique} called BYOP to enable the effective generation of radiomic features, which is necessary as the true bounding boxes are often absent. BYOP leverages an interpretable learning technique to supply estimated bounding boxes dynamically during training.
\item Excellent experimental \textbf{results} achieved on the NIH Chest X-ray benchmark \cite{wang2017chestx}, using very few annotations. Besides improving the disease classification AUC from 82.8\% to 83.8\%, our framework significantly boosts the localization results, by an average of 2\% over different IoU thresholds, compared to reported baselines. Figure \ref{fig:teaser} provides a visualization example showing our localization results to be more robust and accurate than the previous results from CheXNet \cite{rajpurkar2017chexnet}, 
\end{itemize}

\section{Related Work}
\textbf{Self-supervised Learning and Contrastive Learning:} Self-supervision uses pre-formulated (or ``pretext") tasks to train with unlabeled data. Popular handcrafted pretext tasks include solving jigsaw puzzles \cite{noroozi2016unsupervised}, relative patch prediction \cite{doersch2015unsupervised} and colorization \cite{zhang2016colorful}. However, many of these tasks rely on ad-hoc heuristics that could limit the generalization and transferability of learned representations. Consequently, contrastive learning of visual representations has emerged as the front-runner for self-supervision and has demonstrated superior performance on downstream tasks \cite{he2020momentum, chen2020improved,chen2020simple, chen2020big,misra2020self,grill2020bootstrap}. Most of those successes take place in the natural image domain due to the ease of creating contrastive views by applying data augmentations.

There have been a few recent attempts towards contrastive learning in \underline{multi-domain knowledge} tasks. \cite{tian2019contrastive} studied knowledge transfer via contrastive learning, including from one sensory domain knowledge to another (e.g., RGB to Depth). \cite{pielawski2020comir} contrasted bright-field and second-harmonic generation microscopy images for their registration: two image domain knowledge captured for biomedical applications, with large appearance discrepancy between them. \cite{zhang2021cross} aimed to learn text-to-image synthesis by maximizing the mutual information between image and text, through multiple contrastive losses which capture inter-domain and intra-domain correspondences. The commodity among those existing works is that their two domains (e.g., image and text, RGB and depth, or two types of microscopy images) are both readily available for training. \textbf{In contrast to our task}, the other radiomic features domain knowledge is not available ahead of training and needs to be adaptively bootstrapped from the image domain during training.

\textbf{Contrastive Learning for Medical Image Analysis:} Self-supervised learning using pre-text tasks has been recently popular in medical image analysis \cite{spitzer2018improving,bai2019self,zhou2019models,zhu2020rubik,holmberg2020self}. When it comes to contrastive learning, \cite{chaitanya2020contrastive} proposed a domain-specific pretraining strategy, by extracting contrastive pairs from MRI and CT datasets using a combination of localized and global loss functions, which relies on the availability of both MRI and CT scans. \cite{zhang2020contrastive} leveraged contrastive learning to infer transferable medical visual representations from paired images and text modalities. 

Prior work adopting contrastive learning on chest X-rays remains scarce, due to the roadblock of creating two contrastive views. \cite{liu2019align} explicitly contrasted X-rays with pathologies against healthy ones using attention networks, in a fully supervised setting. The closest recent work to our knowledge \cite{han2021pneumonia} also applied radiomic features to guide contrastive learning for detecting pneumonia
in chest X-rays. However, their method needs to apply a pre-trained ResNet on the images to generate attention for radiomic features extraction and therefore relied on a multi-stage training heuristic. Their method hence produced no joint optimization of two domain knowledge, and lacked the localization capability. 


\textbf{Radiomics in Medical Diagnosis:} Radiomics studies have demonstrated their power in image-based biomarkers for cancer staging and prognostication \cite{nasief2019machine}. Radiomics extracts quantitative data from medical images to represent tumor phenotypes, such as spatial heterogeneity of a tumor and spatial response variations. \cite{eilaghi2017ct} demonstrated that radiomic of CT texture features are associated with the overall survivalrate of pancreatic cancer. \cite{chen2017assessment} revealed that the first-order radiomic features (e.g., mean, skewness, and kurtosis) are correlated with pathological responses to cancer treatment. \cite{huang2018added} showed that radiomics could increase the positive predictive value and reduce the false-positive rate in lung cancer screening for small nodules compared with human reading by thoracic radiologists. \cite{zhang2018learning} found that multiparametric MRI-based radiomics nomograms provided improved prognostic ability in advanced nasopharyngeal carcinoma (NPC). In comparison, deep learning algorithms are often criticized for being ``black box" and lack interpretability despite high predictive accuracy. That limitation has motivated many interpretable learning techniques including activation maximization \cite{erhan2009visualizing}, network inversion \cite{mahendran2015understanding}, GradCAM \cite{selvaraju2016grad}, and network dissection \cite{bau2017network}. We believe that the joint utilization of radiomics and interpretable learning techniques in our framework can further advance accurate yet interpretable learning in the medical image domain.


\section{Method}

\begin{figure*}[]
\centering
\includegraphics[width=\linewidth]{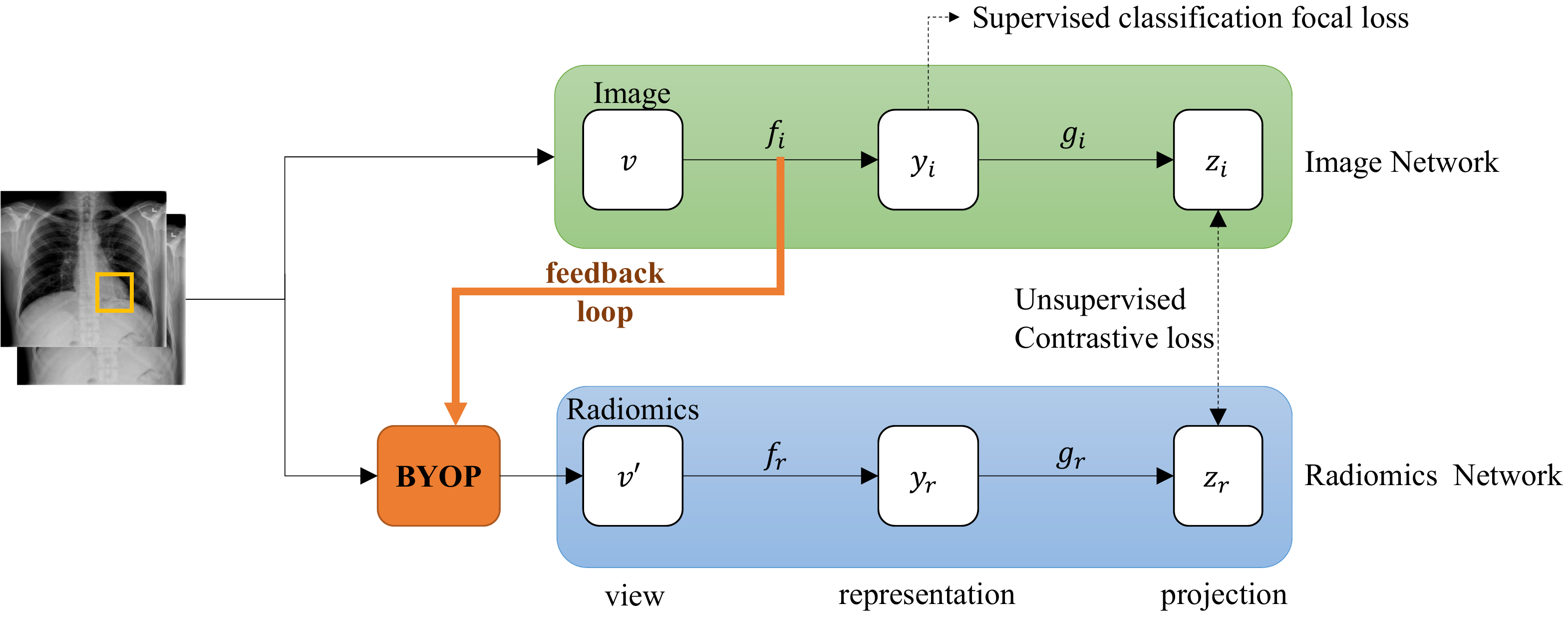}
  \caption{Overview of our proposed framework. 
  During training, given a set of images, very few images have annotations, our framework provides two views: the image and the radiomic features (generated by the BYOP module, the detail view is shown in Figure \ref{fig:byop}). From the image view $v$, we output a representation $y_i=f_i (v)$ and a projection $z_i=g_i (y_i)$ via an image encoder $f_i$ and image projection head $g_i$, respectively. Similarly, from the radiomic view $v'$, we output $y_r=f_r (v')$ and the radiomic projection $z_r=g_r (y_r)$ via a radiomic encoder $f_r$ and radiomic projection head $g_r$, respectively. We maximize agreement between $z_i$ and $z_r$ via a contrastive loss (NT-Xnet). In addition, we minimize the classification errors from representation $y_i$ via a focal loss. During testing, only the image encoder is kept and applied to the new X-rays.
  }
\label{fig:model}
\end{figure*}

\textbf{The Framework.}
Our goal is to learn an image representation $y_i$ which can then be used for disease classification and localization. Our framework uses two neural networks to learn: the image and radiomics networks. The image network consists of an encoder $f_i$ (ResNet-18) and a projector $g_i$ (two-layer MLPs with ReLU). The radiomics network has a similar architecture as the image network, but uses another three-layer MLPs for radiomic encoder $f_r$ and a different set of weights for the projector $g_r$. The proposed architecture is summarized in Figure \ref{fig:model}.





The primary innovation of our method lies in how we select positive and negative examples, which will be expanded below in Section \ref{findPN} and Section \ref{BYOP}. We also formulate the semi-supervised loss for our problem when a small amount of annotated data is available in Section \ref{baseline}. The entire framework can be trained from end to end, and the representation $y_i$ will be used for downstream disease classification and localization tasks.



\subsection{Finding Positive and Negative Samples: Data-Driven Learning Meets Domain Expertise}

\label{findPN}
The reasons to use contrastive learning as our framework are three-fold. First, contrastive learning leverages unlabeled data and we have few disease localization (bounding boxes) annotations available. Second, empirical findings \cite{yang2020rethinking,kangexploring} prove that contrastive learning is robust in classification tasks with class-imbalanced datasets. In clinical settings, most medical image datasets suffer an extreme class-imbalance problem \cite{gao2020handling}. 
Third, contrastive learning naturally fits ``multi-view" concepts. In our case, we are still comparing two different views of the same subject, but unlike classic contrastive learning where two views are from the same domain space, our views for positive sampling are from different domain knowledge (\cite{ye2018hierarchical} proved that views from multi-domain knowledge should also align), while our negative sampling is from the same domain knowledge. In the subsequent section, we will describe our unique positive and negative sampling methodologies in more detail.

\vspace{.5em}
\noindent
\textbf{Positive Sampling.}
To obtain a positive pair of views, we randomly select an image labeled with a given disease and generate two views for it. The first view will be its image features and another view will be its radiomic features.
We decided to leverage radiomic features for the second view as traditional image augmentation strategies cannot be leveraged here.
Furthermore, radiomic features have labels, are naturally more interpretable than the image features extracted from deep learning-based image encoders.

Obtaining the radiomic features for our dataset is a ``chicken-and-egg" problem. \textbf{Radiomic features are highly sensitive and dependent on local regions} for which we do not have local bounding box annotations. Meanwhile, we need to make the image features similar to the radiomic features to learn from radiomic features to better learn localization of the abnormalities. This process means that \textbf{bounding boxes generation is dependent on radiomic features} which forms a loop cycle.
To address this issue, we design the \textit{\underline{B}ootstrap \underline{Y}our \underline{O}wn \underline{P}ositive Samples} (\textbf{BYOP}) method using such a feedback module. For more details, see Section \ref{BYOP}.

\begin{figure}
\centering
\includegraphics[width=1\linewidth]{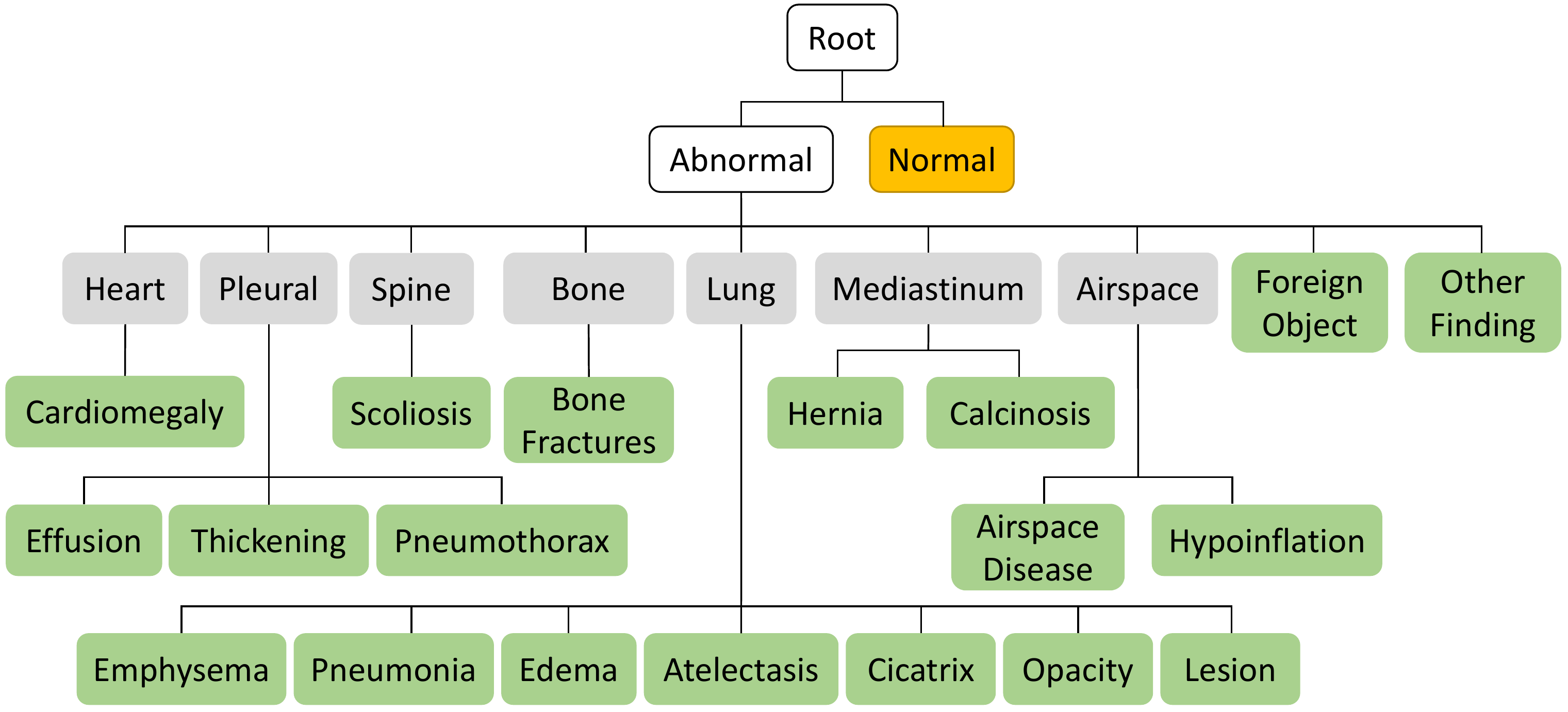}
  \caption{Disease hierarchy relationship predefined based on domain expertise, reprinted from~\cite{zhang2020radiology}. 
  }
\label{fig:disease}
\end{figure}

\vspace{.5em}
\noindent
\textbf{Negative Sampling.}
The original images are used for views of the negative samples because the same domain is supposed to be more similar and thus harder for the model to distinguish between the positive and negative samples, leading to a more robust model \cite{robinson2020contrastive}. 
Besides, the image features focuses on local regions highlighted by the attention map rather than the whole image.
To identify harder negative samples, we go one step further, by not only selecting any random image, but ``hard similar" images. 
Here, we first get prior knowledge from the pre-constructed disease hierarchy relationship for image negative sampling, shown in Figure \ref{fig:disease}, defined by \cite{zhang2020radiology}. The pre-constructed disease hierarchy relationship is initialed with 21 nodes. In this hierarchy, each disease (green) belongs to a body part (grey). We therefore only treat normal chest X-rays or images within the same body part but with a different disease as negative examples. We call these negative examples  “hard similar” images in this study. As an example, if our ``anchor" image is labeled as “Pneumonia/Lung” , our “hard similar” images should include “Atelectasis/Lung”, “Edema/Lung”, or “Normal” but not “Bone Fractures”.

\subsection{Bootstrap Your Own Positive Samples (BYOP) with Radiomics in the Feedback Loop}
\label{BYOP}

\begin{figure}
\centering
\includegraphics[width=\linewidth]{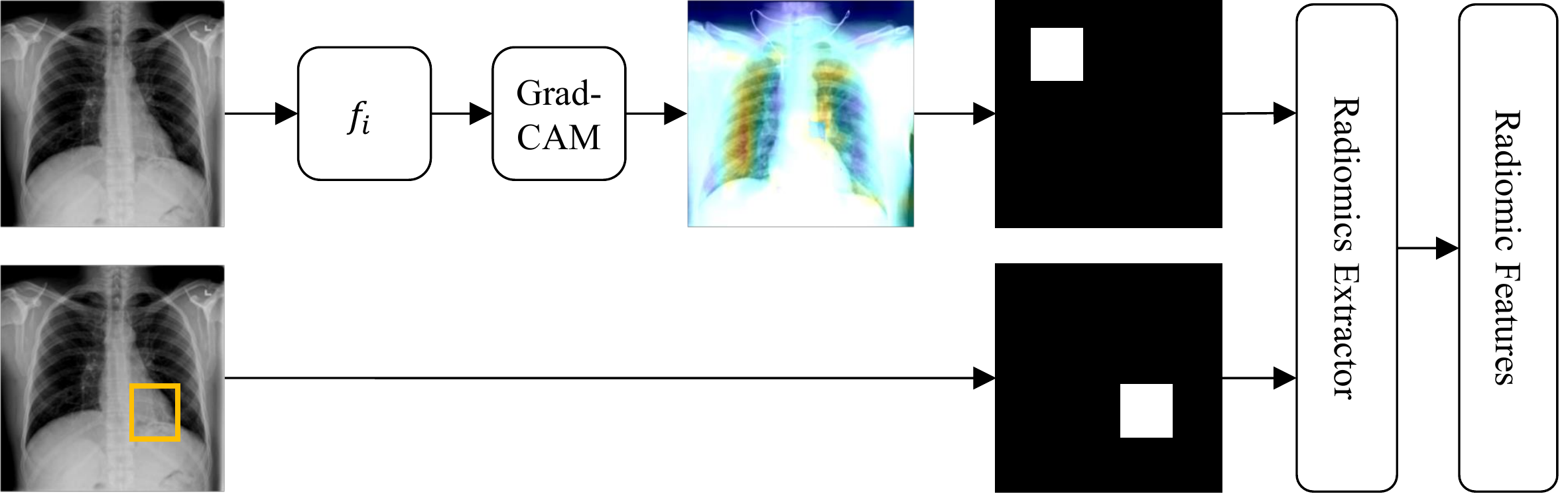}
  \caption{Overview of our \textit{BYOP module}. 
  For the unannotated images, we leverage \textit{Grad-CAM} to generate heatmaps and apply an ad-hoc threshold to generate the bounding boxes. For the annotated images, we directly use the ground-truth bounding boxes. Then with the combination of generated bounding boxes and ground-truth bounding boxes, we use the \textit{Pyradiomic} tool as the radiomic extractor to extract the radiomic features. Note that the generated radiomic features are the combination of the accurate and `pseudo' radiomic features for annotated and unannotated images, respectively.
  }
\label{fig:byop}
\end{figure}

The core component of our cross-modal contrastive learning is the \textit{Bootstrap Your Own Positive Samples} (\textbf{BYOP}) module. BYOP leverages a feedback loop to learn region localization from generated radiomic features as the positive sample for the image features. The architecture of BYOP is shown in Figure \ref{fig:byop}.
The BYOP contains two components, bounding box generation and radiomic features extraction.

\vspace{.5em}
\noindent
\textbf{Bounding Boxes Generation.} 
We feed the fourth layer of the image encoder $f_i$ (i.e., ResNet-18) to the Gradient-weighted Class Activate Mapping (Grad-CAM) \cite{selvaraju2016grad} to extract attention maps and apply an ad-hoc threshold to generate bounding boxes from the attention maps. 

\vspace{.5em}
\noindent
\textbf{Radiomic Features Extraction.}
The radiomic features are composed of the following categories:
\begin{itemize}
    \item First-Order statistics features measure the distribution of voxel intensities within the bounding boxes. The features include energy (the measurement of the magnitude of voxel values), entropy (the measurement of uncertainty in the image values), and max/mean/median gray level intensity within the region of interest (ROI), etc. 
    \item Shape-based features include features like Mesh Surface, Pixel Surface, Perimeter, and etc.
    \item Gray-level features include a gray-level features include a Gray Level Co-occurance Matrix (GLCM) features, a Gray Level Size Zone (GLSZM) features, a Gray Level Run Length Matrix (GLRLM) features, a Neighboring Gray Tone Difference Matrix (NGTDM) features, and a Gray Level Dependence Matrix (GLDM) features. 
\end{itemize}
Given the original images and generated bounding boxes, we used the Pyradiomic tool to extract radiomic features \cite{van2017computational}.

\subsection{Semi-Supervised Loss Function}
\label{baseline}
Our framework is mixed with supervised classification and unsupervised contrastive learning. 
For the localization task, we use the knowledge-augmented contrastive loss for unsupervised contrastive learning. 
For the classification task, we could have used standard cross-entropy loss, but considering that the chest X-ray dataset is highly imbalanced, we instead find focal loss more helpful \cite{pasupa2020convolutional}.
We briefly review the two loss functions below.

\vspace{.5em}
\noindent
\textbf{Unsupervised Knowledge-Augmented Contrastive Loss.}
Our cross-modal contrastive loss function extends the normalized temperature-scaled cross-entropy loss (NT-Xent). We randomly sample a minibatch of $N$ examples and define the contrastive prediction task on pairs of augmented examples derived from the minibatch. Let $v_{bd}$ be the image in the minibatch with disease $d$ and body part $b$, and $\operatorname{sim}(u,v)$ be the cosine similarity. The loss function $\ell_{v_{bd}}$ for a positive pair of example $(v_{bd},v_{bd}')$ is defined as
\begin{equation*}
\ell_{v_{bd}}=-\log \frac{\exp \left(\operatorname{sim}\left(\boldsymbol{z}_{i}(v_{bd}), \boldsymbol{z}_{r}(v_{bd}^\prime)\right) / \tau\right)}{\sum \mathbbm{1}_{[k = b, l \neq d]} \exp \left(\operatorname{sim}\left(\boldsymbol{z}_{i}(v_{bd}), \boldsymbol{z}_{i}(v_{kd})\right) / \tau\right)}
\end{equation*}
where $\mathbbm{1}_{[k = b, l \neq d]} \in \{0, 1\}$ is an indicator function evaluating to 1 iff $k = b$ and $l \neq d$. $\tau$ is the temperature parameter. The final unsupervised contrastive loss $\mathcal{L}_{cl}$ is computed across all disease-positive images in the minibatch.




\vspace{.5em}
\noindent
\textbf{Supervised Focal Loss.}
We feed the output of the image encoder $f_i$ to a simple linear classifier. 
The supervised classification focal loss is defined as 
\begin{equation*}
    \mathcal{L}_{f l}=\left\{\begin{array}{ll}
-\alpha\left(1-y^{\prime}\right)^{\gamma} \log y^{\prime}, & y=1 \\
-(1-\alpha) y^{\prime \gamma} \log \left(1-y^{\prime}\right), & y=0
\end{array}\right.
\end{equation*}
$\alpha$ allows us to give different importance to positive and negative examples. $\gamma$ is used to distinguish easy and hard samples and force the model to learn more from difficult examples. 

Eventually, we treat it as multi-task learning  (one task is supervised disease classification and one is unsupervised contrastive learning) and the total loss is defined as 
\begin{equation*}
    \mathcal{L}= \lambda \times \mathcal{L}_{cl}+ (1 -\lambda)\times \mathcal{L}_{fl} 
\end{equation*}

\section{Experiments}

\begin{table*}[tbh]
\centering
\small
\begin{spacing}{1}
\begin{tabular}{@{}l*{9}{c@{~~}}c@{}}
\toprule
Method & Atelectasis  & Cardiomegaly  & Effusion  & Infiltration  & Mass  & Nodule  & Pneumonia  & Pneumothorax  & \textbf{Mean}\\ 
\midrule
\textit{Wang et. al.}\cite{wang2017chestx} &  0.72 & 0.81 & 0.78 & 0.61 & 0.71 & 0.67 & 0.63 & 0.81 & 0.718\\
\textit{Wang et. al.}\cite{wang2018tienet} &  0.73  & 0.84  & 0.79    & 0.67  & 0.73  &  0.69 &  0.72  &  0.85 & 0.753\\
\textit{Yao et. al.}\cite{yao2017learning} &  0.77 & 0.90 & 0.86 & 0.70 & 0.79 & 0.72 & 0.71 & 0.84 & 0.786\\
\textit{Rajpurkar et. al.}\cite{rajpurkar2017chexnet} & 0.82 & 0.91 & 0.88 & 0.72 & 0.86 & 0.78 & 0.76 & 0.89 & 0.828\\
\textit{Kumar et. al.}\cite{kumar2018boosted} & 0.76 & 0.91 & 0.86 & 0.69 & 0.75 & 0.67 & 0.72 & 0.86 & 0.778\\ 
\textit{Liu et. al.}\cite{liu2019align} & 0.79 & 0.87 & 0.88 & 0.69 & 0.81 & 0.73 & 0.75 & 0.89 & 0.801\\
\textit{Seyyed et. al.}\cite{seyyed2020chexclusion} & 0.81 & 0.92 & 0.87 & 0.72 & 0.83 & 0.78 & 0.76 & 0.88 & 0.821\\ \midrule
Our model             & \textcolor{red}{0.84} & \textcolor{red}{0.93} & \textcolor{red}{0.88} & \textcolor{red}{0.72} & \textcolor{red}{0.87} & \textcolor{red}{0.79} & \textcolor{red}{0.77} & \textcolor{red}{0.90} & \textcolor{red}{0.838}\\\bottomrule
\end{tabular}
\caption{Comparison with the baseline models for AUC of each class and average AUC. For each column, \textcolor{red}{red} values denote the best results.
}
\label{AUC}
\end{spacing}
\vspace{-1em}
\end{table*}

\textbf{Dataset and Protocol Setting.} We evaluated our framework using the NIH Chest X-ray dataset \cite{wang2017chestx}. 
It contains 112,120 X-ray images collected from 30,805 patients. As other large chest X-ray datasets, this dataset is also extremely class imbalanced: the healthy cases (84,321 front-view images) are far more than cases with diseases (24,624 front-view images), and different disease occurrence frequencies vary dramatically. The disease labels were extracted from radiology reports with a rule-based tool~\cite{peng2018negbio}. There are 9 classes, specifically one for ``No findings" and 8 for diseases (Atelectasis, Cardiomegaly, Effusion, Infiltration, Mass, Nodule, Pneumonia, and Pneumothorax).
The disease labels are expected to have above 90\% accuracy. In addition, the dataset includes 984 bounding boxes for 8 types of chest diseases annotated for 880 images by radiologists. We separate the images with provided bounding boxes from the entire dataset. Hence, we have two sets of images called ``annotated" (880 images) and ``unannotated" (111,240 images).

In our experiment, we follow the same protocol of \cite{wang2017chestx}, to shuffle the unannotated dataset into three subsets: $70\%$ for training, $10\%$ for validation, and $20\%$ for testing. For the annotated dataset, we randomly split the dataset into two subsets: $20\%$ for training and $80\%$ for testing. Note that there is no patient overlap between all the sets.

\textbf{Evaluation Metrics.}
For the disease classification task, we use Area under the Receiver Operating Characteristic curve (AUC) to measure the performance of our model. 
For the disease localization task, we evaluate the detected regions against annotated ground truth bounding boxes, using intersection over union ratio (IoU). The localization results are only calculated on the test set of the annotated dataset. The localization is defined as correct only if IoU $>$ T(IoU), where T(*) is the threshold.

\textbf{Implementation Details.}
We use the ResNet-18 model as the image encoder. We initialize the image encoder with the weights from the pre-trained ImageNet model except for the last fully-connected layer. We set the batch size as 64 and train the model for 30 epochs. We optimize the model by the Adam method and decay the learning rate by 0.1 from 0.001 every 5 epochs. Furthermore, we use linear warmup for the first 10 epochs only for the disease classification task, which helps the model converge faster to generate stable heatmaps. We train our model on AWS with one Nvidia Tesla V100 GPU. The model is implemented in PyTorch.


\subsection{Disease Classification}


Table \ref{AUC} shows the AUC of each class and a mean AUC across the 8 chest diseases. Compared to a series of relevant baseline models, our proposed model achieves better AUC scores for the majority of diseases. The overall improvement in performance is remarkable when compared to other models except CheXNet \cite{rajpurkar2017chexnet}. One possible reason for our lack of improvement can be that \cite{rajpurkar2017chexnet}'s backbone is DenseNet-121, which is much deeper than the ResNet-18 in our model. It thus, able to capture much more discriminative features than our ResNet-18. Despite the fact, our model still achieves better or comparable results than CheXNet, which demonstrates that the cross-modal contrastive learning branch boosts the robustness of the image features without the need to increase the complexity of the backbone. Specifically, the performance of our model demonstrates significant improvements for disease abnormalities with larger associated regions on the image, such as ``Ateclectasis", ``Cardiomegaly", and ``Pneumothorax". In addition, small objects features like ``Mass" and ``Nodule", are recognized as well as in CheXNet. In summary, these experimental results show the superiority of our proposed model over relevant other methodologies.

\subsection{Disease Localization}

\begin{table*}[tbh]
\centering
\small
\begin{tabular}{@{}l@{~~}l*{9}{c@{~~~~}}c@{}}
\toprule
T(IoU) & Model & Atelectasis & Cardiomegaly & Effusion & Infiltration & Mass & Nodule & Pneumonia & Pneumothorax & \textbf{Mean}\\
\midrule
0.1 & \textit{Wang et. al.}\cite{wang2017chestx} & 0.69 & 0.94 & 0.66 & 0.71 & 0.40 & 0.14 & 0.63 & 0.38 & 0.569\\
 & \textit{Li et. al.}\cite{li2018thoracic} & 0.71 & \textcolor{red}{0.98} & 0.87 & 0.92 & 0.71 & 0.40 & 0.60 & 0.63 & 0.728\\
 & Our model & \textcolor{red}{0.72} & 0.96 & \textcolor{red}{0.88} & \textcolor{red}{0.93} & \textcolor{red}{0.74} & \textcolor{red}{0.45} & \textcolor{red}{0.65} & \textcolor{red}{0.64} & \textcolor{red}{0.746}\\
\midrule
0.2 & \textit{Wang et. al.}\cite{wang2017chestx} & 0.47 & 0.68 & 0.45 & 0.48 & 0.26 & 0.05 & 0.35 & 0.23 & 0.371\\
 & \textit{Li et. al.}\cite{li2018thoracic} & 0.53 & \textcolor{red}{0.97} & 0.76 & 0.83 & 0.59 & 0.29 & 0.50 & 0.51 & 0.622\\
 & Our model & \textcolor{red}{0.55} & 0.89 & \textcolor{red}{0.78} & \textcolor{red}{0.85} & \textcolor{red}{0.62} & \textcolor{red}{0.31} & \textcolor{red}{0.52} & \textcolor{red}{0.54} & \textcolor{red}{0.633}\\
\midrule
0.3 & \textit{Wang et. al.}\cite{wang2017chestx} & 0.24 & 0.46 & 0.30 & 0.28 & 0.15 & 0.04 & 0.17 & 0.13 & 0.221\\
 & \textit{Li et. al.}\cite{li2018thoracic} & 0.36 & \textcolor{red}{0.94} & 0.56 & 0.66 & 0.45 & 0.17 & 0.39 & 0.44 & 0.496\\
 & Our model & \textcolor{red}{0.39} & 0.85 & \textcolor{red}{0.60} & \textcolor{red}{0.67} & 0.43 & \textcolor{red}{0.21} & \textcolor{red}{0.40} & \textcolor{red}{0.45} & \textcolor{red}{0.500}\\
\midrule
0.4 & \textit{Wang et. al.}\cite{wang2017chestx} & 0.09 & 0.28 & 0.20 & 0.12 & 0.07 & 0.01 & 0.08 & 0.07 & 0.115\\
 & \textit{Li et. al.}\cite{li2018thoracic} & \textcolor{red}{0.25} & \textcolor{red}{0.88} & 0.37 & 0.50 & 0.33 & 0.11 & 0.26 & 0.29 & 0.374\\
 & Our model & 0.24 & 0.81 & \textcolor{red}{0.42} & \textcolor{red}{0.54} & \textcolor{red}{0.34} & \textcolor{red}{0.13} & \textcolor{red}{0.28} & \textcolor{red}{0.32} & \textcolor{red}{0.385}\\
\midrule
0.5 & \textit{Wang et. al.}\cite{wang2017chestx} & 0.05 & 0.18 & 0.11 & 0.07 & 0.01 & 0.01 & 0.03 & 0.03 & 0.061\\
 & \textit{Li et. al.}\cite{li2018thoracic} & 0.14 & \textcolor{red}{0.84} & 0.22 & 0.30 & 0.22 & 0.07 & \textcolor{red}{0.17} & 0.19 & 0.269\\
 & Our model & 0.16 & 0.77 & \textcolor{red}{0.29} & 0.35 & 0.24 & \textcolor{red}{0.09} & 0.15 & \textcolor{red}{0.22} & \textcolor{red}{0.284}\\
\midrule
0.6 & \textit{Wang et. al.}\cite{wang2017chestx} & 0.02 & 0.08 & 0.05 & 0.02 & 0.00 & 0.01 & 0.02 & 0.03 & 0.029\\
 & \textit{Li et. al.}\cite{li2018thoracic} & 0.07 & 0.73 & 0.15 & \textcolor{red}{0.18} & 0.16 & 0.03 & 0.10 & 0.12 & 0.193\\
 & Our model & \textcolor{red}{0.09} & \textcolor{red}{0.74} & \textcolor{red}{0.19} & 0.16 & \textcolor{red}{0.18} & \textcolor{red}{0.04} & \textcolor{red}{0.11} & \textcolor{red}{0.14} & \textcolor{red}{0.206}\\
\midrule
0.7 & \textit{Wang et. al.}\cite{wang2017chestx} & 0.01 & 0.03 & 0.02 & 0.00 & 0.00 & 0.00 & 0.01 & 0.02 & 0.011\\ 
 & \textit{Li et. al.}\cite{li2018thoracic} & 0.04 & 0.52 & 0.07 & 0.09 & 0.11 & 0.01 & 0.05 & 0.05 & 0.118\\
 & Our model & 0.05 & \textcolor{red}{0.54} & \textcolor{red}{0.09} & 0.11 & 0.12 & \textcolor{red}{0.02} & \textcolor{red}{0.07} & 0.06 & \textcolor{red}{0.133}\\
\bottomrule
\end{tabular}
\caption{Disease localization accuracy comparison under different IoU thresholds. \textcolor{red}{Red} numbers denote the best result for each column. 
}
\label{tab:disease loc}
\end{table*}

\begin{table*}[tbh]

\centering
\small
\begin{spacing}{1}
\begin{tabular}{@{}l*{8}{c@{~~~}}c@{}}
\toprule
Method & Atelectasis  & Cardiomegaly  & Effusion  & Infiltration  & Mass  & Nodule  & Pneumonia  & Pneumothorax  & \textbf{Mean}\\ 
\midrule
Base & 0.75 & 0.85 & 0.83 &  0.67 & 0.69 & 0.64 & 0.70 & 0.79 & 0.740\\
w. FL & 0.78 & 0.84 & 0.80 & 0.68 & 0.76 & 0.72 & 0.72 & 0.82 & 0.765\\
w. BYOP & 0.82 & 0.90 & 0.85 & 0.71 & 0.82 & 0.75 & 0.74 & 0.86 & 0.806\\
Full model             & \textcolor{red}{0.84} & \textcolor{red}{0.93} & \textcolor{red}{0.88} & \textcolor{red}{0.72} & \textcolor{red}{0.87} & \textcolor{red}{0.79} & \textcolor{red}{0.77} & \textcolor{red}{0.90} & \textcolor{red}{0.838}\\\bottomrule
\end{tabular}
\caption{Ablation studies on focal loss and BYOP module for disease classification. \textcolor{red}{Red} numbers denote the best result for each column.}
\label{discussion}
\end{spacing}
\vspace{-1em}
\end{table*}

We compare our disease localization accuracy to other state-of-the-art models under different IoU thresholds (Table \ref{tab:disease loc}). Since disease localization is not an easy task in chest X-ray images, we did not find as many other methods as for disease classification task. To our knowledge, we only have two baseline methods from \cite{wang2017chestx} and \cite{li2018thoracic}. From these comparisons, we find our model significantly outperforms baselines by an average of 2\% over different IoU thresholds. Importantly, our model is able to perform well not only on the easier tasks, but also for more difficult ones like localizing “Mass” and “Nodule”, where the disease localization is within a small area. When the IoU threshold is set to 0.1, our model outperforms others on all diseases except for ``Cardiomegaly". As the IoU threshold increases, our framework is superior to other models in terms of better accuracy and maintains this superior performance. For instance, when the threshold increases, the IoUs of ``Cardiomegaly" decrease less than the baselines and even outperform the baselines when IoU threshold is above 0.5.

We prefer a higher IoU threshold, specifically, IoU = 0.7, for disease localization because we expect high-accuracy disease localization application is necessary for clinical applications. To this end, the method we propose is superior to the baseline by a slight margin.

It is also worth nothing that, for some diseases, such as Pneumonia and Infiltration, the localization of disease can appear in multiple places while only one bounding box is provided for each image. Hence, it is reasonable that our model does not align well with the ground truth when the threshold is as small as 0.1, especially for Pneumonia and Infiltration.
Overall, our model outperforms the reference models for almost all IoU thresholds.

\begin{figure*}
\begin{center}
\includegraphics[width=1\linewidth]{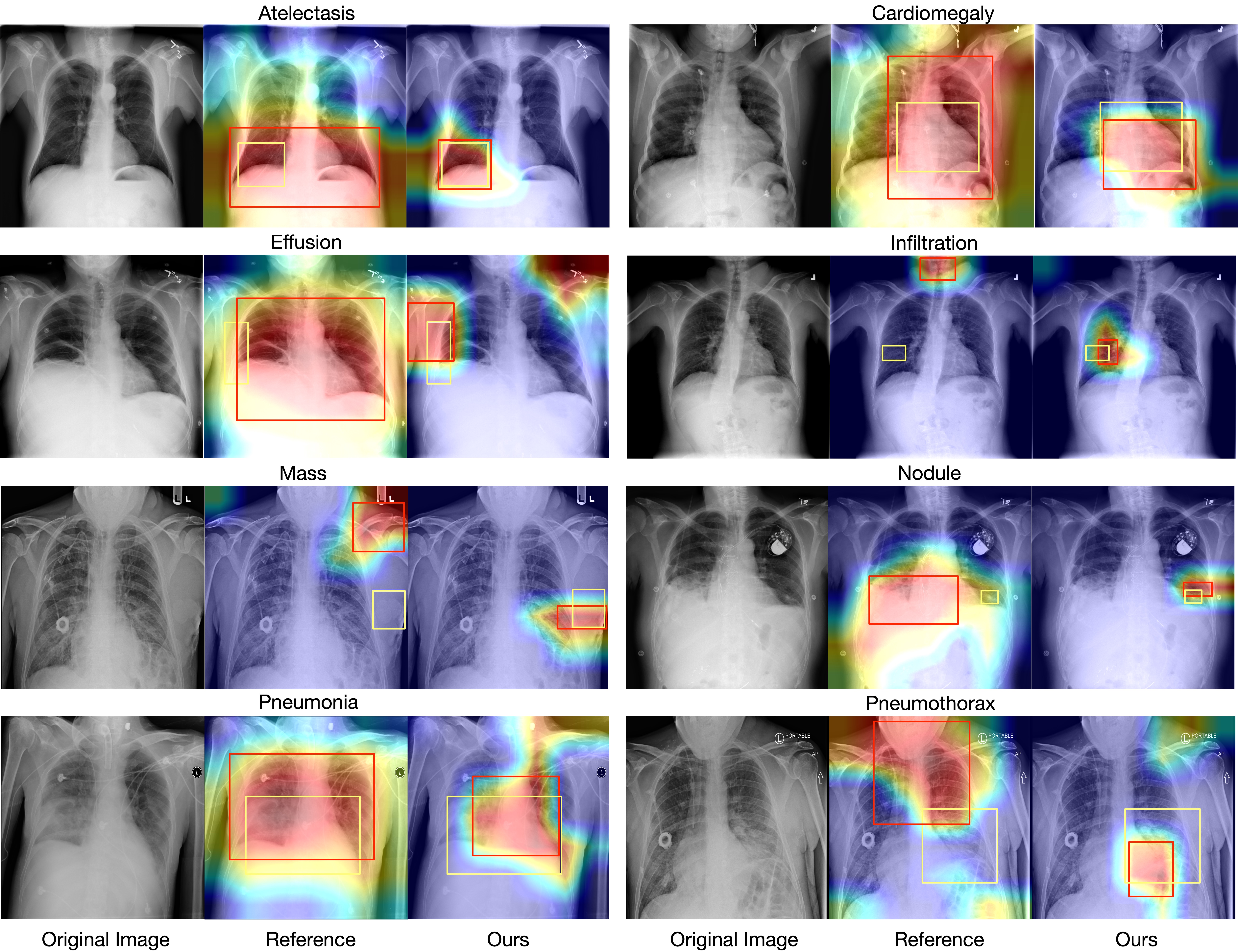}
\end{center}
\caption{Examples of visualization of localization on the test images. We plot the results of diseases near thoracic. The attention maps are generated from the fourth layer of ResNet-18. The ground-truth bounding boxes and the predicted bounding boxes are shown in yellow and red, respectively. The left image in each pair is the original chest X-ray image, the middle one is the localization result of CheXNet \cite{rajpurkar2017chexnet} and the right one is our localization result. All examples are positive for corresponding disease labels. Best viewed in color. }
\label{fig:attention}
\end{figure*}

\subsection{Ablation Discussion}
In this section, we study the contribution of our BYOP module on both disease classification and localization tasks. 

\textbf{Disease Classification.} For this task, note that the use of focal loss should also boost the model with the class-imbalanced chest X-ray dataset. Thus, we compare the performance of our base model with only focal loss (labeled ``w. FL") or with only the BYOP module (labeled ``w. BYOP"), respectively. As shown in Table \ref{discussion}, although both focal loss and BYOP improve the model performance, BYOP contributed more strongly. This stronger contribution is expected since BYOP tends to generate more robust radiomic features, which further reinforces the image encoder to focus on the image region that contains the targeted disease.

\textbf{Disease Localization.} Note that our base model is a ResNet-18 image encoder, which is not as powerful as CheXNet \cite{rajpurkar2017chexnet} with DenseNet-121. Thus we compare the performance of our model with CheXNet. As shown in Figure \ref{fig:attention}, our localization result is superior to the CheXNet. For the example of `Atelectasis', `Cardiomegaly', `Effusion', `Nodule', `Pneumonia' and `Pneumothorax', while the baseline model tends to focus on a large area of the image, our model precisely captures the correct disease location. For harder localization cases like `Mass' and `Nodule', the baseline model's focus is incorrect and does not have any overlap with the ground-truth areas while our model still predicts perfectly. The results demonstrate that the BYOP module significantly boosts the model performance.


\section{Conclusion}
In this work, we propose a semi-supervised, end-to-end knowledge-augmented contrastive learning model that can jointly model disease classification and localization with limited localization annotation data. Our approach differs from previous studies in the choice of data augmentation, the use of radiomic features as prior knowledge, and a feedback loop for image and radiomic features to mutually reinforce each other. Additionally, the project aims to address current gaps in radiology by making prior knowledge more accessible to image data analytic and diagnostic assisting tools, with the hope that this will increase the model's interpretability. Experimental results demonstrate that our method outperforms the state-of-the-art algorithms, especially for the disease localization task, where our method can generate more accurate bounding boxes. Importantly, we hope the method developed here is inspiring for the future research on incorporating different kinds of prior knowledge of medical images with contrastive learning.

\section*{Acknowledgement}
This project is funded by Amazon Machine Learning Grant and NSF AI Center at UT Austin. It also was supported by the National Library of Medicine under Award No. 4R00LM013001.

{\small
\bibliographystyle{ieee_fullname}
\bibliography{egbib}
}

\end{document}